\documentclass{article}
\usepackage{spconf,amsmath,graphicx}
\usepackage[caption=false,font=footnotesize]{subfig} % 用 \subfloat 就用它
\usepackage{booktabs}
\usepackage{xcolor}
\usepackage{balance}
\usepackage{cite}
\usepackage[hidelinks]{hyperref} % 尽量最后加载

\title{DELNet: Continuous All-in-One Weather Removal via Dynamic Expert Library}
\name{\textit{Shihong Liu}$^{1}$, \textit{Kun Zuo}$^{2}$, \textit{Hanguang Xiao}$^{1*}$}

\address{
$^{1}$School of Artificial Intelligence, Chongqing University of Technology, Chongqing 401135, China\\
$^{2}$Sun Yat-sen University, Shenzhen Campus, Shenzhen, Guangdong 518107, China\\
52232313112@stu.cqut.edu.cn, zuok@mail2.sysu.edu.cn\\ simenxiao1211@163.com\\
* Corresponding author
}

\begin{document}
%\ninept
%
\maketitle
\begin{abstract}
All-in-one weather image restoration methods are valuable in practice but depend on pre-collected data and require retraining for unseen degradations, leading to high cost. We propose DELNet, a continual learning framework for weather image restoration. DELNet integrates a judging valve that measures task similarity to distinguish new from known tasks, and a dynamic expert library that stores experts trained on different degradations. For new tasks, the valve selects top-k experts for knowledge transfer while adding new experts to capture task-specific features; for known tasks, the corresponding experts are directly reused. This design enables continuous optimization without retraining existing models. Experiments on OTS, Rain100H, and Snow100K demonstrate that DELNet surpasses state-of-the-art continual learning methods, achieving PSNR gains of 16\%, 11\%, and 12\%, respectively. These results highlight the effectiveness, robustness, and efficiency of DELNet, which reduces retraining cost and enables practical deployment in real-world scenarios.
\end{abstract}

\setlength{\parskip}{0pt}  % 段落间距为 0
\setlength{\parsep}{0pt}
\setlength{\topsep}{0pt}
\setlength{\partopsep}{0pt}
\setlength{\textfloatsep}{4pt}     % 图表与正文间距
\setlength{\intextsep}{4pt}        % 插图插入文本的间距
\setlength{\floatsep}{3pt}         % 图表之间的间距
% 紧凑排版（放在 \usepackage{spconf,...} 之后）
\setlength{\parskip}{0pt}
\setlength{\textfloatsep}{4pt plus 1pt minus 1pt}  % 图/表与正文
\setlength{\floatsep}{4pt plus 1pt minus 1pt}      % 浮动体之间
\setlength{\intextsep}{4pt plus 1pt minus 1pt}     % 文中浮动与正文
\setlength{\abovecaptionskip}{2pt}
\setlength{\belowcaptionskip}{0pt}
\setlength{\abovedisplayskip}{2pt}
\setlength{\belowdisplayskip}{2pt}
\setlength{\textfloatsep}{6pt plus 1pt minus 1pt}
\setlength{\abovecaptionskip}{2pt}
\setlength{\belowcaptionskip}{0pt}

\section{Introduction}Adverse weather restoration has evolved from single-task models (rain, fog, snow, low-light, etc.) to all-in-one networks \cite{SUN2026111875,yi2025multi,wen2025multi,10446531}. However, existing methods either rely on multiple encoders–decoders with high switching cost or unified structures that overlook degradation differences, both lacking continual learning. This limits deployment in dynamic environments such as autonomous driving \cite{xiong2023neural}.

Continual learning \cite{10445917} enables adaptation to new tasks but suffers from catastrophic forgetting. Memory-based solutions \cite{cheng2024continual} partially alleviate this issue but introduce high overhead and limited feature diversity.

\begin{figure}[!t]
	\centering
	\includegraphics[width=.95\columnwidth]{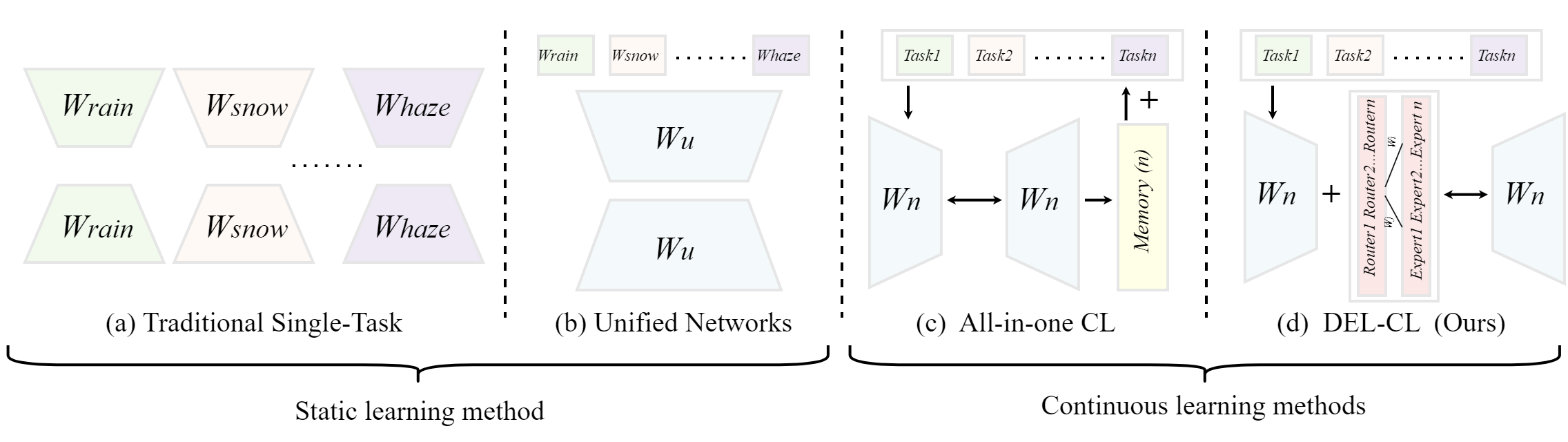}
	\caption{Comparison chart of methods for eliminating adverse weather conditions. Including single removal method, unified method, all-in-one continuous learning method and our method.}
	\label{FIG:1}
\end{figure}

To overcome these challenges, we propose the Dynamic Expert Library Network (DELNet) (Fig.~\ref{FIG:1} d), which integrates a judging valve to identify task similarity, a dynamic expert library to preserve old knowledge while adding new experts, and a feature enhancement backbone to improve restoration quality. This design achieves continual adaptation without forgetting, enabling robust multi-weather image restoration.

\section{Methods}

\subsection{Overall architecture}
\begin{figure}[]
	\centering
	\includegraphics[width=.9\columnwidth]{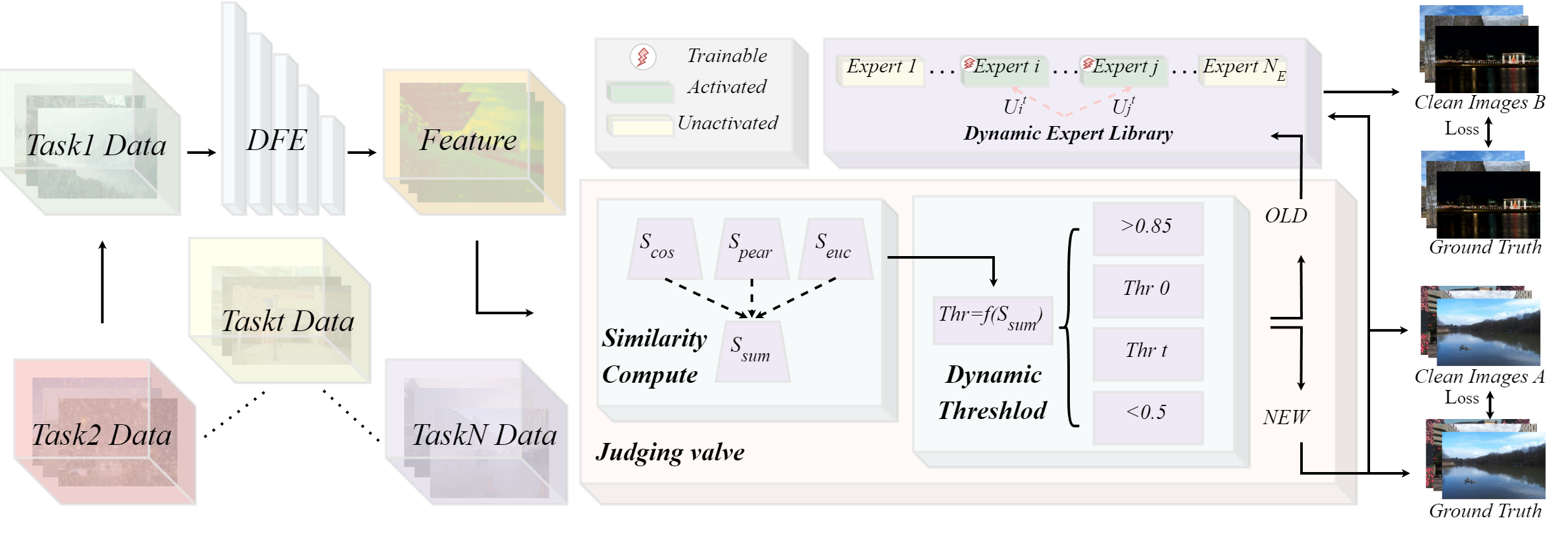}
	\caption{Overview of DELNet. The framework integrates a Deep Feature Enhancement (DFE) network, a judging valve, and a dynamic expert library for continual multi-weather image restoration.}
	\label{fig2}
\end{figure}

The overall architecture of DELNet (Fig.~\ref{fig2}) consists of a Deep Feature Enhancement (DFE) network, a judging valve, and a dynamic expert library. The DFE employs parallel self-attention and polarization attention with soft maximization and weight mapping to strengthen channel–spatial features for high-quality restoration. The judging valve generates task feature vectors, compares them with historical tasks, and classifies each as new or old. For new tasks, top-k experts are selected and updated with new adapters; for old tasks, corresponding experts are reused. Finally, outputs of active experts are fused, performance scores are updated, and trained experts are frozen to prevent forgetting, enabling continual learning across diverse weather degradations.

\subsection{Judging Valve}
Typical MoE adapters \cite{chen2023mod,gao2024clip} require manually defined task representations to activate routers. To address this limitation, we propose the Judging Valve (JV), which automatically identifies and classifies input tasks without manual intervention.

\textbf{Feature Vectors and Similarity.} For each task $Task_t$, features from the backbone are summarized into a task vector $T^t$ using five statistics: mean, standard deviation, maximum, minimum, and $L_2$ norm. Task similarity is then computed by a combined metric $S_{sum}$, integrating cosine similarity $S_{cos}$, Euclidean similarity $S_{euc}$, and Pearson similarity $S_{pear}$. This enables DELNet to decide whether the current task is new or old. The process is formulated as:

\begin{equation}
S_{cos}(T^1,T^2) = \frac{T^1 \cdot T^2}{\lVert T^1\rVert_2 \lVert T^2\rVert_2},
\end{equation}

\begin{equation}
\lVert T^t\rVert_2 = \left(\sum_{k=1}^d (t_k^t)^2 \right)^{1/2},
\end{equation}

where $T^1 \cdot T^2 = \sum_{k=1}^d t_k^1 t_k^2$ is the dot product of two vectors, $t_k^t$ is the $k$-th element of $T^t$, and $d$ is the vector dimension. 

The Euclidean similarity is obtained from the distance
\begin{equation}
d_{euc} = \lVert T^1 - T^2\rVert_2, \quad 
S_{euc} = \frac{1}{1+d_{euc}},
\end{equation}

while the Pearson similarity $S_{pear}$ measures the linear correlation between task vectors. Finally, the overall similarity is expressed as
\begin{equation}
S_{sum} = a \cdot S_{cos} + b \cdot S_{euc} + c \cdot S_{pear},
\end{equation}
with weights $a=0.5$, $b=0.3$, and $c=0.2$.

 \textbf{Dynamic threshold management.} To enhance adaptability, DELNet employs a dynamic threshold mechanism based on historical task similarities. The first task is initialized as new with similarity 1.0 and threshold 0.75. For later tasks, similarity $>0.85$ indicates old, $<0.5$ indicates new, while intermediate cases trigger threshold updates once samples exceed three. The update is defined as:
\begin{equation}
    Thr_0=MS_{sum}-0.25\times STS_{sum},
\end{equation}
where $MS_{sum}$ and $STS_{sum}$ are the median and standard deviation of similarities. The decision boundary is then adaptively updated:
\begin{equation}
    Thr_{t+1}=Thr_t+e\times clip(Thr_0,{-}f,f),
\end{equation}
with $e=0.05$ controlling update speed and $f=0.05$ bounding changes for stability. In practice, we further include skewness and kurtosis statistics to improve task separability, and bound the dynamic threshold within [0.65, 0.90] for stability.

\subsection{Dynamic Expert Library}
 \textbf{Definition of Dynamic Expert Library.} The Dynamic Expert Library (DEL) is a scalable MoE-based architecture \cite{zhou2022mixture} designed to mitigate catastrophic forgetting. Each expert is implemented as a lightweight adapter composed of instance normalization, projection convolutions, activation, and residual connection. Every expert maintains an independent parameter set and evaluation metric, tailored for specific degradations such as rain, snow, or fog. Formally, the expert set is defined as:
\begin{equation}
    \left\{\varepsilon_{i}\right\}_{i=1}^{N_E},
\end{equation}
where $N_E$ is the number of experts (30 in our implementation, adjustable by task scale).

\begin{figure}
	\centering
	\includegraphics[width=.8\columnwidth]{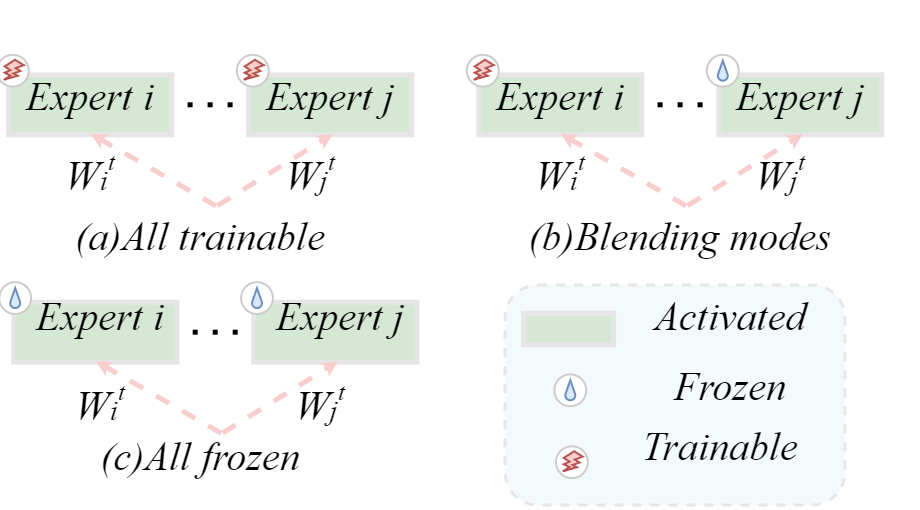}
	\caption{Three different adapter and expert modes: (a) All trainable, (b) Blending modes, and (c) All frozen.}
	\label{fig3}
\end{figure}

\textbf{Expert scheduling selection.} DEL employs a performance–usage joint score with Top-$K$ selection to activate only the most relevant experts, while freezing the others (Fig.~\ref{fig3}). For a new task, each expert is assigned a score:
 \begin{equation}
     S_c = P_i \times U_i=P_i \times \frac{1}{C_i + \epsilon},
 \end{equation}
where $P_i$ is the performance score, $C_i$ the usage frequency, and $\epsilon=10^{-6}$ ensures stability. $P_i$ is updated by an exponential moving average:
\begin{equation}
    P^{(t+1)}(i) = \beta_t \cdot P^{t}(i) + (1-\beta_t) \cdot \frac{1}{L_i + \epsilon}.
\end{equation}
with $\beta_t=0.9$ and $L_i$ the task loss.  

The weights of selected experts are computed by loss-based temperature scaling:
\begin{equation}
    u_i = \frac{\exp(-L_i/\tau)}{\sum_{j=1}^{K} \exp(-L_j/\tau)}.
\end{equation}
where $\tau=0.1$ and $K$ is the number of active experts. The final output is the weighted fusion:
\begin{equation}
    y=\sum_{i=1}^{K} u_i \cdot E_i(x),
\end{equation}
with $E_i(x)$ denoting the transformation of expert $i$ on input $x$.

\subsection{Loss function combination strategy}
We design a multi-level objective to promote knowledge transfer and retention: 
(i) reconstruction and contrastive losses for pixel fidelity and semantic consistency; 
(ii) output-level distillation and feature-level projection to mitigate forgetting; and 
(iii) adapter regularization with a dynamic coefficient, complemented by a lightweight expert-diversity term.

\textbf{Reconstruction \& contrastive.}
\begin{equation}
  L_{sw}^{(0)}=\|I_{pred}-I_{gt}\|_{1},
\end{equation}
\begin{equation}
  L_{c}=\mathrm{ContrastLoss}\!\left(I_{pred},\, I_{gt},\, I_{input}\right),
\end{equation}
\begin{equation}
  L_{sw}=L_{sw}^{(0)}+\beta_{1}\,L_{c}.
\end{equation}

\textbf{Distillation (output-level).}
\begin{equation}
  L_{kd}^{(0)}=\|I_{pred\_old}-I_{pred\_new}\|_{1},
\end{equation}
\begin{equation}
  L_{kd}=L_{kd}^{(0)}+\beta_{2}\,\mathrm{ContrastLoss}
  \!\left(I_{pred\_new},\, I_{pred\_old},\, I_{old}\right).
\end{equation}

\textbf{Projection (feature-level).}
\begin{equation}
\begin{aligned}
  h_{old} &= \mathrm{AutoEncoder.pjt}(f_{old}),\\
  h_{new} &= \mathrm{AutoEncoder.pjt}(f_{new}),\\
  L_{p}   &= \|h_{old}-h_{new}\|_{1}.
\end{aligned}
\end{equation}

\textbf{Regularization (adapters).}
\begin{equation}
  L_{reg}=\sum_{i=1}^{N}\|\omega_{i}\|_{2},\qquad
  \beta=0.01\cdot \min\!\left(\tfrac{step}{steps\cdot 5},\, 0.1\right).
\end{equation}

\textbf{Diversity (experts).}
\begin{equation}
  L_{div}=-\gamma \cdot \mathrm{Std}\{L_i\}_{i=1}^{K},\qquad \gamma=0.01,
\end{equation}
where $L_i$ is the loss of the $i$-th active expert among $K$ selected experts. 
This term encourages specialization and reduces redundancy with negligible overhead.

\textbf{Total.}
\begin{equation}
  L_{total}=L_{sw}+\alpha L_{kd}+\lambda L_{p}+\beta L_{reg}+L_{div},
\end{equation}

\noindent\textit{Notation.} 
$\,I_{pred}$: model output; $I_{gt}$: ground truth; $I_{input}$: degraded input.
$I_{pred\_old}$/$I_{pred\_new}$: outputs of old (teacher) / current model on the same old data $I_{old}$.
$f_{old}$/$f_{new}$: encoder features; $h_{old}$/$h_{new}$: their projected features.
$\omega_i$: weight of the $i$-th adapter’s projection layer; $N$: number of adapters.
$\beta_1,\beta_2$: contrastive weights; $\beta$: dynamic regularization weight. We set $\alpha=0.8$ and $\lambda=0.3$ empirically.

\section{Experiment}

\subsection{Datasets and Settings}
We evaluate on RESIDE (haze), Rain100H (rain), and Snow100K (snow) following the order haze$\rightarrow$rain$\rightarrow$snow. 
Training uses Adam ($lr=2\!\times\!10^{-4}$, cosine decay) for $5\times10^5$ iterations, batch size 2, crop size $240\!\times\!240$. 
Evaluation adopts PSNR and SSIM on official test sets. 
All reported results are from the full training setup; a lightweight debug configuration is provided in code but not used for tables.

\begin{figure*}[t]
\centering
\subfloat[RESIDE]{\includegraphics[width=0.32\textwidth]{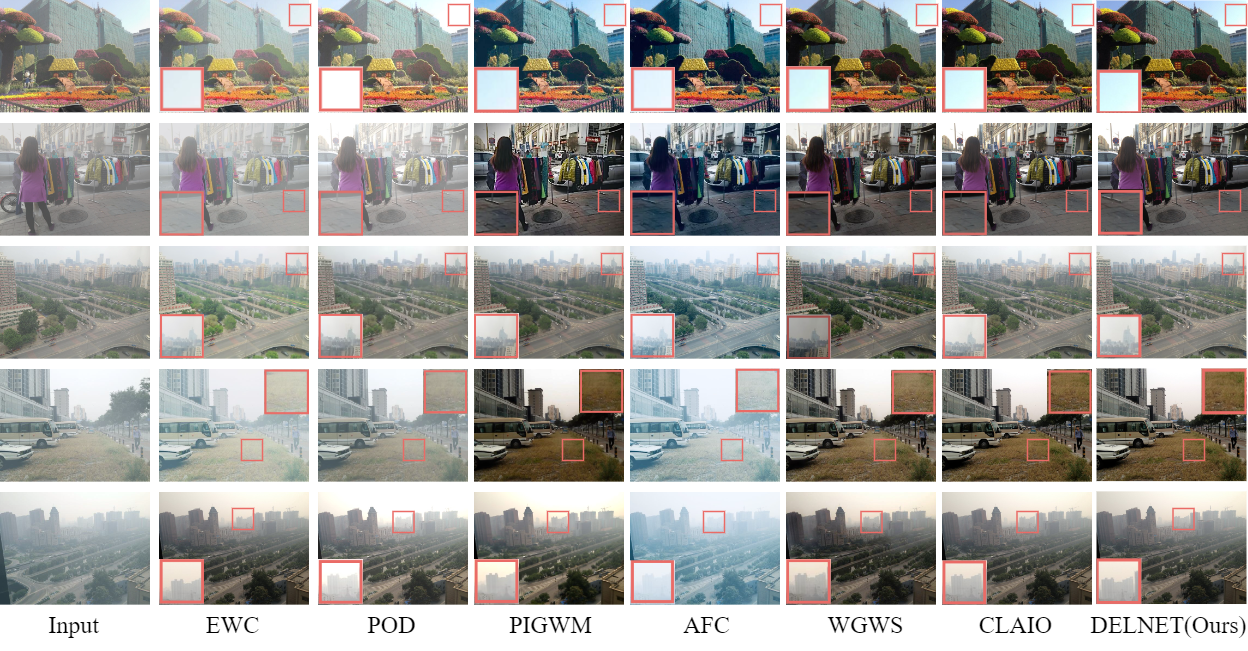}}
\hfill
\subfloat[Rain100H]{\includegraphics[width=0.32\textwidth]{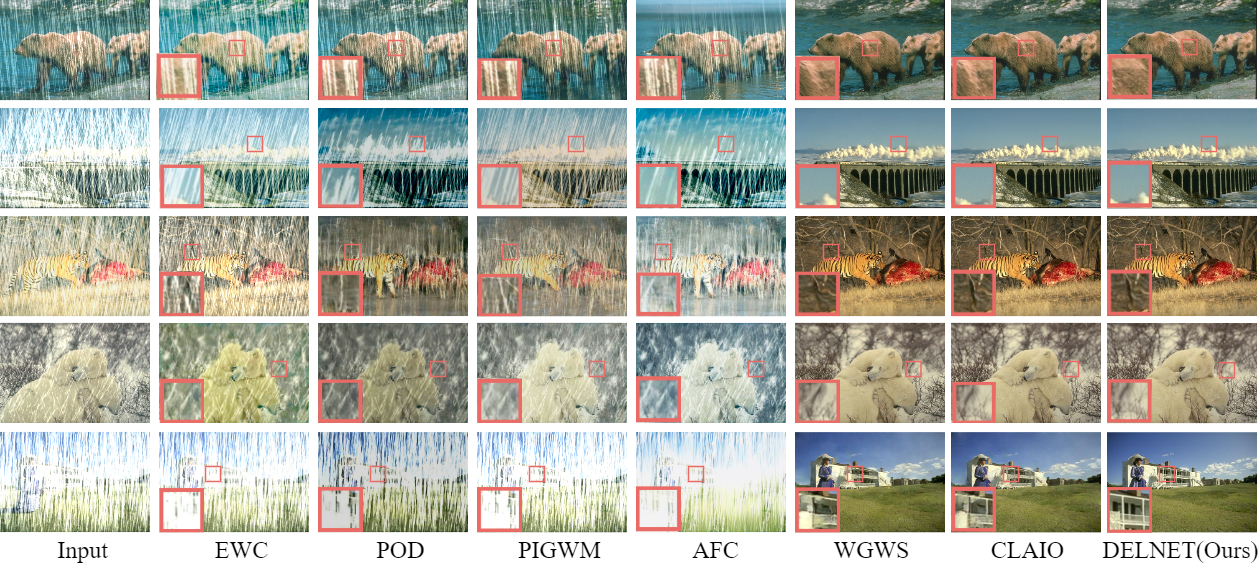}}
\hfill
\subfloat[Snow100K]{\includegraphics[width=0.32\textwidth]{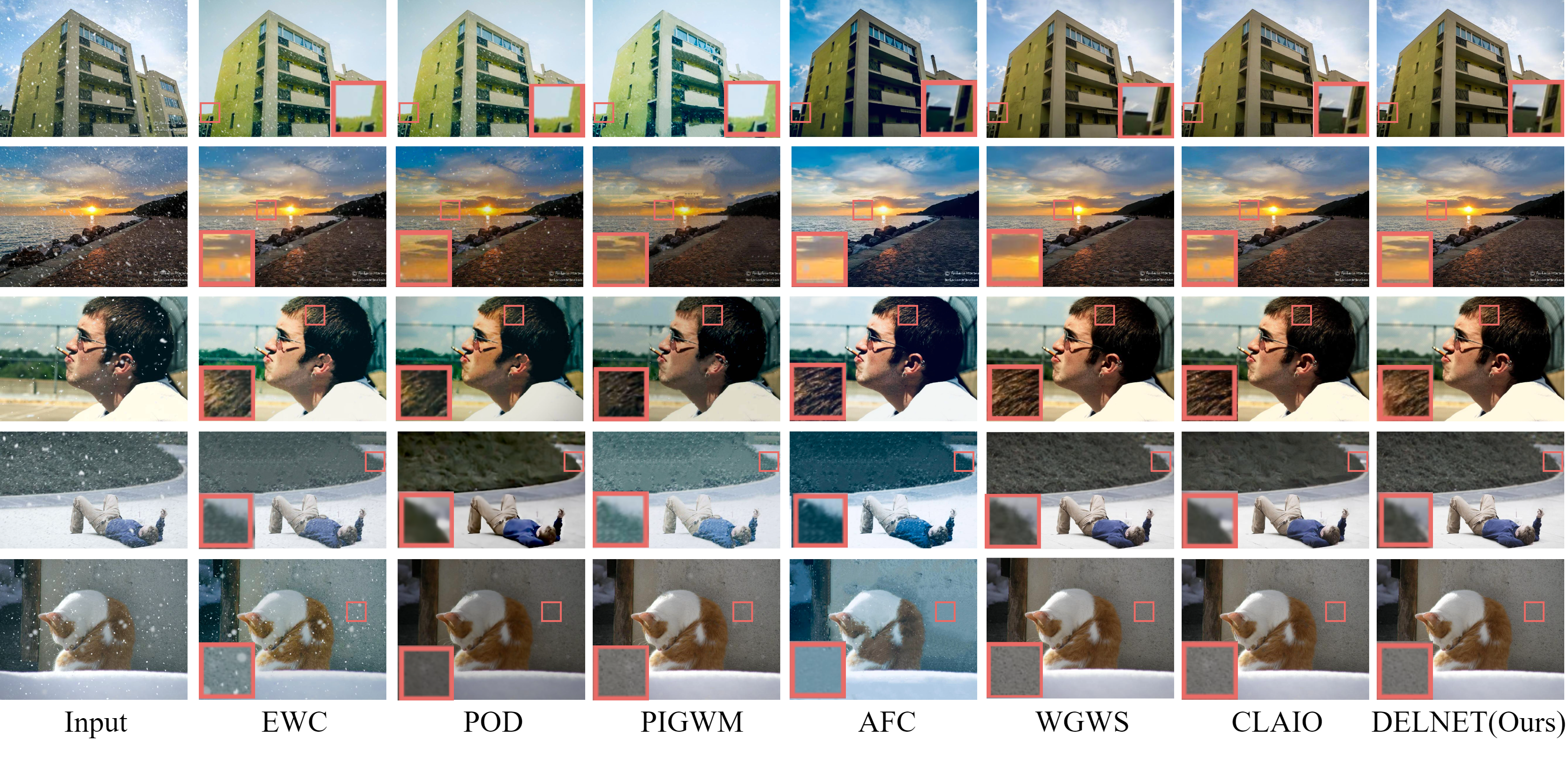}}
\caption{Visualization of image restoration results on RESIDE, Rain100H, and Snow100K datasets.}
\label{figv}
\end{figure*}

\begin{table}[t]
\centering
\scriptsize
\setlength{\tabcolsep}{3pt}
\renewcommand{\arraystretch}{0.95}
\begin{tabular}{lcccc}
\toprule
Methods & Avg & OTS & Rain100H & Snow100K \\
\midrule
EWC \cite{kirkpatrick2017overcoming}     & \textcolor{blue}{27.79/0.81} & 28.23/0.86 & \textcolor{blue}{25.99/0.76} & 29.16/0.80 \\
MAS \cite{aljundi2018memory}    & 25.74/0.73 & \textcolor{blue}{29.11/0.94} & 20.68/0.51 & 27.45/0.73 \\
LwF \cite{li2017learning}    & 20.90/0.64 & 23.12/0.86 & 15.24/0.36 & 24.35/0.71 \\
POD \cite{douillard2020podnet}    & 22.80/0.66 & 23.19/0.82 & 16.56/0.40 & 28.65/0.77 \\
PIGWM \cite{zhou2021image}  & 23.31/0.71 & 28.33/\textcolor{blue}{0.94} & 15.16/0.48 & 26.45/0.73 \\
AFC \cite{kang2022class}    & 25.73/0.77 & 26.73/0.83 & 20.48/0.66 & \textcolor{blue}{29.98/0.82} \\
DELNet (Ours) & \textcolor{red}{31.28/0.94} & \textcolor{red}{31.22/0.98} & \textcolor{red}{29.03/0.89} & \textcolor{red}{33.58/0.94} \\
\midrule
TransWeather \cite{valanarasu2022transweather} & 27.82/0.88 & 27.12/0.91 & 28.53/0.88 & 27.83/0.85 \\
WGWS \cite{zhu2023learning}         & 29.61/0.91 & 29.54/0.94 & \textcolor{blue}{29.10/0.88} & 30.19/0.90 \\
Chen et al. \cite{chen2022learning}  & 27.66/0.87 & 27.34/0.90 & 28.69/0.87 & 26.97/0.82 \\
AIRFormer \cite{10196308}   & 26.04/0.85 & 25.67/0.95 & 24.46/0.74 & 28.01/0.85 \\
WeatherDiff \cite{10021824}  & 29.94/0.90 & 24.30/0.95 & 26.66/0.85 & 28.86/0.89 \\
ADSM \cite{10855564}        & 30.21/0.91 & 30.66/0.92 & \textcolor{red}{29.74/0.89} & 30.24/0.91 \\
CLAIO \cite{cheng2024continual}        & \textcolor{blue}{30.67/0.93} & \textcolor{blue}{31.05/0.98} & 28.66/0.88 & \textcolor{blue}{32.31/0.93} \\
DELNet (Ours) & \textcolor{red}{31.28/0.94} & \textcolor{red}{31.22/0.98} & 29.03/\textcolor{blue}{0.89} & \textcolor{red}{33.58/0.94} \\
\bottomrule
\end{tabular}
\caption{Haze, rain, and snow removal in the single-task sequence. Top: continual learning methods. Bottom: integrated methods. Results in PSNR/SSIM, best in red, second best in blue.}
\label{tab1}
\end{table}

\begin{table}[t]
\centering
\scriptsize
\setlength{\tabcolsep}{3pt}
\renewcommand{\arraystretch}{0.95}
\begin{tabular}{lcccc}
\toprule
Methods & Avg & OTS & Rain100H & Snow100K \\
\midrule
EWC \cite{kirkpatrick2017overcoming}       & 24.61/0.71 & 27.69/0.78 & 20.69/0.63 & 25.47/0.73 \\
MAS \cite{aljundi2018memory}      & 24.74/0.79 & 26.38/0.90 & 23.15/0.69 & 24.71/0.77 \\
LwF \cite{li2017learning}      & 21.41/0.70 & 21.85/0.86 & 16.99/0.42 & 25.40/0.81 \\
POD \cite{douillard2020podnet}      & 23.60/0.78 & 23.66/0.83 & 18.35/0.70 & \textcolor{blue}{28.81/0.81} \\
PIGWM \cite{zhou2021image}    & 24.44/\textcolor{blue}{0.87} & \textcolor{blue}{28.93}/\textcolor{red}{0.95} & 17.44/0.81 & 26.95/\textcolor{blue}{0.85} \\
AFC \cite{kang2022class}      & \textcolor{blue}{26.96}/0.83 & 27.94/0.82 & \textcolor{blue}{24.84}/\textcolor{blue}{0.87} & 28.10/0.81 \\
DELNet   & \textcolor{red}{29.69}/\textcolor{red}{0.91} & \textcolor{red}{30.06}/\textcolor{blue}{0.94} & \textcolor{red}{28.74}/\textcolor{red}{0.88} & \textcolor{red}{30.29}/\textcolor{red}{0.91} \\
\midrule
TransWeather \cite{valanarasu2022transweather} & 27.82/0.88 & 27.12/0.91 & 28.53/0.88 & 27.83/0.85 \\
WGWS \cite{zhu2023learning}         & 29.61/\textcolor{blue}{0.91} & 29.54/0.94 & \textcolor{blue}{29.10}/0.88 & 30.19/0.90 \\
Chen et al. \cite{chen2022learning}  & 27.66/0.87 & 27.34/0.90 & 28.69/0.87 & 26.97/0.82 \\
AIRFormer \cite{10196308}   & 26.04/0.85 & 25.67/\textcolor{blue}{0.95} & 24.46/0.74 & 28.01/0.85 \\
WeatherDiff \cite{10021824}  & \textcolor{blue}{29.94}/0.90 & 24.30/\textcolor{red}{0.95} & 26.66/0.85 & 28.86/0.89 \\
ADSM \cite{10855564}         & \textcolor{red}{30.21}/0.91 & \textcolor{red}{30.66}/0.92 & \textcolor{red}{29.74}/\textcolor{red}{0.89} & \textcolor{blue}{30.24}/\textcolor{blue}{0.91} \\
CLAIO \cite{cheng2024continual}        & 29.54/0.90 & \textcolor{blue}{30.47}/0.93 & 28.11/0.87 & 30.05/0.90 \\
DELNet       & 29.69/\textcolor{red}{0.91} & 30.06/0.94 & 28.74/\textcolor{blue}{0.88} & \textcolor{red}{30.29}/\textcolor{red}{0.91} \\
\bottomrule
\end{tabular}
\caption{Results of haze, rain, and snow removal using multi-task sequence (others same as Table~\ref{tab1})}
\label{tab2}
\end{table}

\begin{table}[t]
\centering
\scriptsize
\setlength{\tabcolsep}{3pt}
\begin{tabular}{lcc}
\toprule
Methods & Outdoor-Rain (PSNR/SSIM) & Params \\
\midrule
All-in-One \cite{li2020allinone}      & 24.71/0.8980 & 44.0M \\
TransWeather \cite{valanarasu2022transweather}   & 28.83/0.9000 & 38.1M \\
AirNet \cite{Liairn_2022_CVPR}         & 25.69/0.8993 & --    \\
Chen et al. \cite{chen2022learning}    & 23.94/0.8500 & 28.7M \\
WGWS \cite{zhu2023learning}           & 25.32/0.9070 & 6.0M  \\
LDR \cite{Yang_2024_CVPR}            & 26.92/0.9120 & --    \\
AIRFormer \cite{10196308}       & 24.52/0.7837 & --    \\
WeatherDiff \cite{10021824}    & 25.78/0.8990 & 83.0M \\
ADSM \cite{10855564}           & \textcolor{blue}{29.52/0.9015} & --    \\
CLAIO \cite{cheng2024continual}          & 26.78/\textcolor{blue}{0.9199} & 8.2M  \\
DELNet (Ours)   & \textcolor{red}{29.66/0.9311}  & \textcolor{red}{5.6M} \\
\bottomrule
\end{tabular}
\caption{Outdoor-Rain dataset (deraining + dehazing). Results reported as PSNR/SSIM (baselines same as in Section 3.2.}
\label{tab3}
\end{table}

\begin{table}[t]
\centering
\scriptsize
\setlength{\tabcolsep}{3pt}
\caption{The melting results of components in the overall network.}
\label{tab4}
\begin{tabular}{lcccccc}
\toprule
Combination & FFA & DFE & JV & DEL & PSNR & SSIM \\
\midrule
Baseline & -- & -- & -- & -- & 13.83 & 0.4267 \\
C1 & $\sqrt{}$ & -- & -- & -- & 16.71 & 0.7914 \\
C2 & -- & $\sqrt{}$ & -- & -- & 17.32 & 0.7974 \\
C3 & -- & $\sqrt{}$ & $\sqrt{}$ & -- & 24.76 & 0.8125 \\
C4 & -- & $\sqrt{}$ & $\sqrt{}$ & $\sqrt{}$ & \textcolor{red}{31.22} & \textcolor{red}{0.9784} \\
\bottomrule
\end{tabular}
\end{table}

% \usepackage{multirow}
\begin{table}[t]
\centering
\scriptsize
\setlength{\tabcolsep}{3pt}
\caption{Experimental results of ablation using different loss functions.}
\label{tab5}
\begin{tabular}{lccccccc}
\toprule
Loss & Lsw & Lkd & Lp & Lreg & Lct & PSNR & SSIM \\
\midrule
C5 & $\sqrt{}$ & -- & -- & -- & -- & 25.97 & 0.8486 \\
C6 & $\sqrt{}$ & $\sqrt{}$ & -- & -- & -- & 28.99 & 0.9072 \\
C7 & $\sqrt{}$ & $\sqrt{}$ & $\sqrt{}$ & -- & -- & 29.97 & 0.9354 \\
C8 & $\sqrt{}$ & $\sqrt{}$ & $\sqrt{}$ & $\sqrt{}$ & -- & 29.99 & 0.9691 \\
C9 & $\sqrt{}$ & $\sqrt{}$ & $\sqrt{}$ & $\sqrt{}$ & $\sqrt{}$ & \textcolor{red}{31.22} & \textcolor{red}{0.9784} \\
\bottomrule
\end{tabular}
\end{table}

% \begin{table}[t]
% \centering
% \scriptsize
% \setlength{\tabcolsep}{4pt}
% \begin{tabular}{lc}
% \toprule
% Experts & Haze (PSNR/SSIM) \\
% \midrule
% 15 & 28.06/0.8467 \\
% 20 & 27.62/0.8142 \\
% 25 & 30.23/0.9325 \\
% 30 & \textcolor{red}{31.22/0.9784} \\
% 35 & 25.96/0.8433 \\
% \bottomrule
% \end{tabular}
% \caption{Ablation on the number of experts in DELNet.}
% \label{tab:experts}
% \end{table}

% \usepackage{multirow}
% \usepackage{multirow}

% \usepackage{multirow}
% \usepackage[normalem]{ulem}

% \usepackage{multirow}

\begin{table}[t]
\centering
\scriptsize
\setlength{\tabcolsep}{2pt}
\begin{tabular}{lcccc}
\toprule
Task Seq. & Avg & OTS & Rain100H & Snow100K \\
\midrule
rain–snow–haze & 29.58/0.93 & 28.19/\textcolor{red}{0.98} & \textcolor{red}{29.65}/0.89 & 30.90/0.91 \\
rain–haze–snow & 29.37/0.93 & 28.11/0.97 & 29.07/\textcolor{blue}{0.89} & 30.94/0.92 \\
haze–snow–rain & 29.77/0.92 & 29.58/0.95 & 28.35/0.87 & 31.40/\textcolor{red}{0.93} \\
haze–rain–snow & 29.69/0.91 & \textcolor{red}{30.06}/0.94 & 28.74/0.88 & 30.29/0.91 \\
snow–rain–haze & \textcolor{blue}{30.21}/\textcolor{blue}{0.93} & 29.45/0.97 & \textcolor{blue}{29.11}/\textcolor{red}{0.89} & \textcolor{red}{32.09}/\textcolor{blue}{0.93} \\
snow–haze–rain & \textcolor{red}{30.26}/\textcolor{red}{0.93} & \textcolor{blue}{29.70}/\textcolor{blue}{0.97} & 29.03/0.89 & \textcolor{blue}{32.06}/0.93 \\
\bottomrule
\end{tabular}
\caption{Continual learning under different task orders (PSNR/SSIM).}
\label{tab6}
\end{table}

\begin{figure}[!t]
\centering
% 左右都加 [t]，保证“顶端对齐”
\subfloat[]{%
\begin{minipage}[t]{0.52\columnwidth}
  \vspace{0pt}% 关键：把顶部基线钉在最上边
  \centering
  \includegraphics[height=0.55\linewidth]{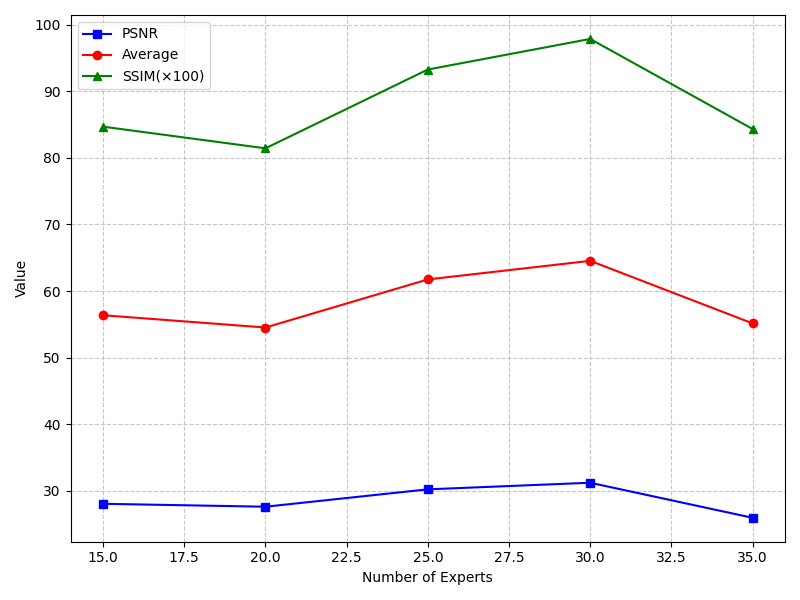}
\end{minipage}
}\hfill
% --- 右：表格（顶端对齐） ---
\subfloat[]{%
\begin{minipage}[t]{0.42\columnwidth}
  \vspace{0pt}% 关键：同样顶端对齐
  \centering
  \scriptsize
  \setlength{\tabcolsep}{3pt}
  \renewcommand{\arraystretch}{0.95}
  % 用 \linewidth 让表格宽度与右侧容器一致；tabular 也设为 [t]
  \resizebox{\linewidth}{!}{%
    \begin{tabular}[t]{lc}
    \toprule
    Experts & Haze (PSNR/SSIM) \\
    \midrule
    15 & 28.06/0.8467 \\
    20 & 27.62/0.8142 \\
    25 & 30.23/0.9325 \\
    30 & \textcolor{red}{31.22/0.9784} \\
    35 & 25.96/0.8433 \\
    \bottomrule
    \end{tabular}
  }% end resizebox
\end{minipage}
}

\vspace{-2mm}
\caption{Ablation on expert number: (a) Impact of the number of experts (b) Ablation on expert number (Haze).}
\label{fig5}
\end{figure}

\subsection{Method comparison}
We evaluate DELNet under two settings: single-task sequence (tasks trained/tested sequentially) and multi-task sequence (tasks jointly trained/tested), following CLAIO \cite{cheng2024continual}. Comparisons include continual learning baselines \cite{kirkpatrick2017overcoming,li2017learning,aljundi2018memory,douillard2020podnet,zhou2021image,kang2022class} and static all-in-one models \cite{valanarasu2022transweather,10021824,zhu2023learning}, using official codes and identical datasets. Results are reported in Table \ref{tab1}, Table \ref{tab2}, and Fig.~\ref{figv}.

DELNet achieves consistent gains: on single-task sequences it improves PSNR by 2.11–3.54dB over continual baselines and up to 1.27dB over static methods, reaching 29.03dB on Rain100H (vs.~29.74dB best). It produces clearer targets with natural lighting while requiring only 5.6M parameters, far fewer than TransWeather (38.1M), WeatherDiff (83.0M), and WGWS (6M) (Table \ref{tab3}). On mixed rain–fog data, it further surpasses the second-best by +0.14dB / +0.0112 SSIM and the strongest static baseline by +4.95dB. These results confirm DELNet’s robustness, efficiency, and adaptability under continual learning.

% \usepackage{multirow}

\subsection{Ablation study}
\begin{figure}[!t]
  \centering
  \subfloat[Input]{%
    \includegraphics[width=0.31\linewidth,keepaspectratio]{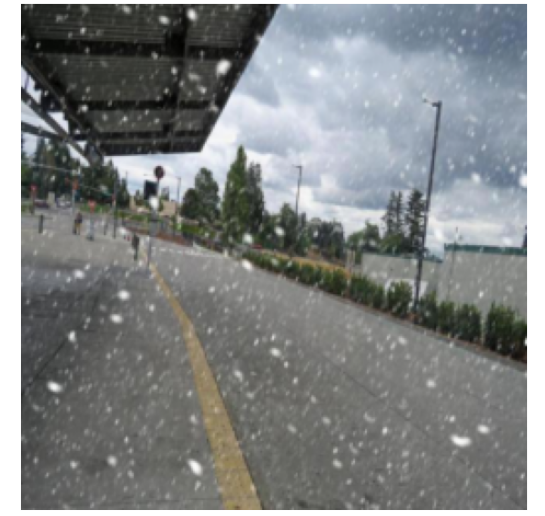}}\hfill
  \subfloat[Similarity]{%
    \includegraphics[width=0.31\linewidth,keepaspectratio]{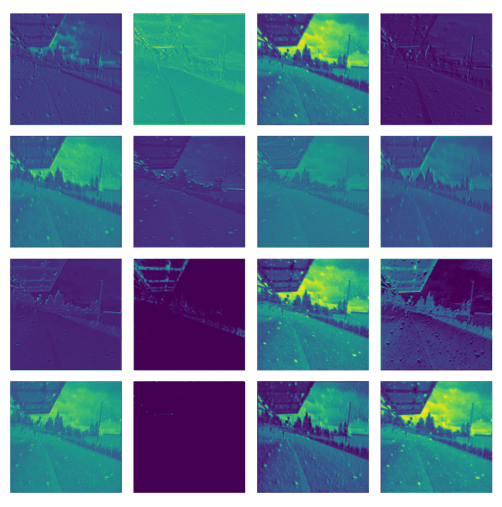}}\hfill
  \subfloat[Difference]{%
    \includegraphics[width=0.31\linewidth,keepaspectratio]{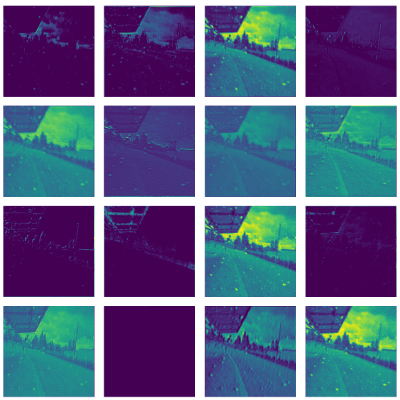}}
  \caption{Visualization of feature maps for extracting intermediate features in rainy weather.}
  \label{net2}
\end{figure}

We conduct ablation studies on the OTS dehazing dataset \cite{li2018benchmarking} to validate the effectiveness of each module and loss design. As shown in Table \ref{tab4}, integrating the judging valve, DFE, and dynamic expert library yields the best performance. Figure \ref{net2} further visualizes feature extraction, where snow patterns are clearly captured and removed. Figure \ref{fig5} analyze the number of experts: increasing experts enhances knowledge capacity and PSNR/SSIM, but also raises training cost; around 30 experts provides the best trade-off. Table \ref{tab5} compares different loss combinations, confirming that multi-level supervision achieves optimal results. Finally, Table \ref{tab6} explores different task sequences, showing that task order significantly affects continual learning, with earlier tasks achieving better performance. This highlights task-order sensitivity as an open challenge for future work.

\section{Conclusion}
We introduced DELNet, a continual learning framework for all-weather restoration. 
With a judging valve, dynamic expert library, and multi-level objective, DELNet mitigates forgetting and adapts to new degradations without retraining. 
Experiments show consistent gains over baselines. 
Future work will focus on efficient teacher updating and robustness under extreme conditions.

\section*{Acknowledgement}

% （可选）再挤一点点致谢和参考文献之间的空白
\vspace{-0.4em}
This work was supported by the National Natural Science Foundation of China (Grant No. 62471075), the Chongqing’s “Tender-Based System” Project Initiative in the Field of Industry and Information Technology (Grant No. YJX202500-1001002), the Natural Science Foundation of Chongqing (Grant No. CSTB2024NSCQ-LZX0080, CSTB2023TI AD-STX0020, CSTB2025NSCQ-LZX0115, CSTB2023NS CQ-LZX0068, CSTB2022NSCQ-MSX0837). This study does not involve any ethical issue.

\begingroup
\raggedbottom
\balance
\footnotesize
\makeatletter
\def\IEEEbibitemsep{0pt plus .2pt}
\makeatother
\setlength{\itemsep}{0pt}
\setlength{\parskip}{0pt}
\setlength{\parsep}{0pt}
\setlength{\topsep}{0pt}
\setlength{\partopsep}{0pt}

\bibliographystyle{IEEEbib}
\bibliography{refs}

\endgroup

\vfill\pagebreak

\end{document}